\documentclass[manuscript]{copernicus}

\usepackage{booktabs}

\newcommand{\modname}{GPTCast}
\newcommand{\tokname}{VQGAN}
\newcommand{\modnamesmall}{GPTCast-8x8}
\newcommand{\modnamelarge}{GPTCast-16x16}

\begin{document}

\title{\emph{\textbf{\modname}}: a weather language model for precipitation nowcasting}


\Author[1][franch@fbk.eu]{Gabriele}{Franch} 
\Author[1]{Elena}{Tomasi}
\Author[1]{Rishabh}{Wanjari}
\Author[2]{Virginia}{Poli}
\Author[2]{Chiara}{Cardinali}
\Author[2]{Pier Paolo}{Alberoni}
\Author[1]{Marco}{Cristoforetti}

\affil[1]{Fondazione Bruno Kessler, Trento, Italy}
\affil[2]{Arpae Emilia-Romagna, Bologna, Italy}




\runningtitle{GPTCast}

\runningauthor{Gabriele Franch et al.}

\received{}
\pubdiscuss{} 
\revised{}
\accepted{}
\published{}


\firstpage{1}

\maketitle

\begin{abstract}
This work introduces \modname{}, a generative deep-learning method for ensemble nowcast of radar-based precipitation, inspired by advancements in large language models (LLMs). We employ a GPT model as a forecaster to learn spatiotemporal precipitation dynamics using tokenized radar images. The tokenizer is based on a Quantized Variational Autoencoder featuring a novel reconstruction loss tailored for the skewed distribution of precipitation that promotes faithful reconstruction of high rainfall rates. The approach produces realistic ensemble forecasts and provides probabilistic outputs with accurate uncertainty estimation. The model is trained without resorting to randomness, all variability is learned solely from the data and exposed by model at inference for ensemble generation. We train and test \modname{} using a 6-year radar dataset over the Emilia-Romagna region in Northern Italy, showing superior results compared to state-of-the-art ensemble extrapolation methods.
\end{abstract}


\introduction[Introduction and prior work]  
Nowcasting —short-term forecasting up to 6 hours— of precipitation is a crucial tool for mitigating water-related hazards\cite{werner2009understanding}. Sudden precipitation can result in landslides and floods, frequently compounded by strong winds, lightning, and hailstorms, which can seriously jeopardize human safety and damage infrastructure. The foundation of very short-term (up to two hours) precipitation nowcasting systems is the application of extrapolation techniques to weather radar reflectivity sequences\cite{nowcasting2030} that ingest current and $n$ previous observations $T_{-n},\ldots,T_{-1},T_0$ with the aim to extrapolate $m$ future time steps $T_1,T_2,\ldots,T_m$. These short-term precipitation forecasts are essential for emergency response when timely released and properly communicated via early warning systems\cite{FirstWMOWWRPSERAWeatherandSocietyConference}.

The main contender to extrapolation techniques are numerical weather prediction (NWP) models, which can be used to forecast the probability and estimate the intensity of precipitation across large regions, but their accuracy is limited at smaller geographical and temporal scales\cite{AStudyontheScaleDependenceofthePredictabilityofPrecipitationPatterns}. Convective precipitation, which produces high rainfall rates and small cells, is especially difficult to forecast correctly for NWP models\cite{UseofNWPforNowcastingConvectivePrecipitationRecentProgressandChallenges}.
For these reasons, operational weather agencies recognize the great value offered by short-term extrapolation forecasts and make heavy use of statistical and, more recently, data-driven models that utilize the most recent weather radar observations for nowcasting\cite{OperationalApplicationofOpticalFlowTechniquestoRadarBasedRainfallNowcasting, PredictabilityofPrecipitationfromContinentalRadarImagesPartIIIOperationalNowcastingImplementationMAPLE}.

Lagrangian extrapolation is the most well-known method for nowcasting precipitation\cite{bellon1978sharp}. It generates motion vectors to forecast the future direction of precipitation systems by applying optical-flow algorithms to a series of radar-derived rain fields. However, this approach becomes less accurate for increasing lead time, particularly in convective situations where precipitation could increase or decrease quickly. Several alternative techniques have been studied to overcome these constraints, like the seamless integration between nowcasting and NWP forecasts\cite{NowPrecip, steps} and the integration of orography data\cite{OrographicGrowthDecay, nora}. Other, more sophisticated nowcasting methods improve the Lagrangian approach by generating ensemble nowcasts and preserving the precipitation field's structure. These sets of multiple forecasts aid in the assessment of forecast uncertainty by presenting multiple future scenarios. The most widespread example of this approach is the Short-Term Ensemble Prediction System (STEPS)\cite{steps, steps2}.

The most recent advancements in nowcasting precipitation have seen the application of data-driven methods and, more prominently, of Deep Neural Networks (DNNs) and Generative AI techniques to enhance forecast accuracy and realism. Deterministic DNNs have been instrumental in predicting the dynamics of precipitation, including its development and dissipation, overcoming one of the major shortcomings of extrapolation methods\cite{convlstm, Agrawal, predrnn, franchtrajgru, RainNet}. However, deterministic models tend to produce less precise forecasts over time due to increasing uncertainty that manifests as a forecast field that smooths progressively with the lead time. Similarly to Lagrangian extrapolation, to overcome this limitation, ensemble deep learning methods have been introduced. Generative methods have significantly improved the generation of realistic precipitation fields beyond deterministic average predictions. The forefront of this technology is embodied in models that employ techniques such as Generative Adversarial Networks (GANs)\cite{nowcastnet, dgmr}, that enable more accurate and detailed precipitation forecasts by learning to mimic real weather patterns closely, and more recently by Latent Diffusion models\cite{ldcast, prediff}, that can not only generate realistic rainfall forecasts but also produce reliable ensembles that can provide accurate uncertainty quantification of future scenarios. Many of these techniques were originally born in the field of computer vision and subsequently adapted to the weather forecasting domain with resounding success\cite{gan, latentdiffusion}.

In this study, we take inspiration from the successful trend of applying Large Language Models (LLMs) architectures\cite{attention, transformers} born in the field of Natural Language Processing (NLP) to other disciplines\cite{vit, swin}, including medium range weather forecasting domain\cite{aifs, atmorep}, intending to transfer this knowledge to the nowcasting domain. To do so, in our work, we follow a strategy that mimics the setup of natural language processing: a tokenization step, where an input tokenizer splits and maps the input to a finite vocabulary, and an autoregressive model trained on the tokens produced by the tokenizer. We show that such an approach produces realistic and reliable ensemble forecasts. Given the different characteristics of our input data compared to LLMs (i.e., spatiotemporal precipitation fields vs. texts or images), our adaptation introduces several novel contributions instrumental to our task.

\section{\modname{} model architecture}
There are two main components of our approach, which we call \modname{}:
\begin{itemize}
    \item \textbf{Spatial tokenizer}: An image compression and discretization model that learns to map patches of the radar image from/to a finite number of possible representations (tokens). The learned codebook of tokens can be used to express a compact representation of any precipitation field. The tokenizer thus has a dual role: learning how to compress and decompress the information in the input image and how to discretize the compressed information (i.e., learn an optimal codebook). 
    \item \textbf{Spatiotemporal forecaster}: A model trained on token sequences to causally learn the evolutionary dynamics of precipitation over space and time. Given a tokenized spatiotemporal context (a compressed precipitation sequence), the model outputs probabilities over the codebook for the next expected token for the context. The output probabilities can be leveraged for ensemble generation.
\end{itemize}
The two components of the model are trained independently in cascade, starting with the tokenizer. The choice of this dual-stage architecture unlocks a number of desirable properties that are instrumental in meeting many requirements of operational meteorological services when adopting a nowcasting system. The two most important characteristics are realistic ensemble generation and accurate uncertainty estimation. Our architecture provides both realistic ensemble generation capabilities and probabilistic output at the spatiotemporal (token) level.

Another notable feature of \modname{} is its fully deterministic architecture, eliminating the need for random inputs during training or inference. This ensures that all model variability is derived solely from the training data distribution. By learning a discretized representation in the tokenizer, the forecaster can output a categorical distribution over vocabulary, modeling a conditional distribution over possible data values. This approach, unlike continuous variable regression, inherently enables probabilistic outputs. In contrast, all other generative deep learning models\cite{dgmr, ldcast, nowcastnet} require random input during training and inference to promote output variability and generate ensemble members.

The baseline architecture of \modname{} is an adaptation of the work of \citeauthor{esser2021taming}, which we repurposed from the task of image generation to the task of precipitation nowcasting by introducing two key modifications:

\begin{itemize}
    \item In the spatial tokenizer (\tokname{}) model, we replace the standard reconstruction loss (MAE) with a specific loss that helps improve the reconstruction of precipitation patterns (Magnitude Weighted Absolute Error, MWAE). Moreover, the new loss also shows a promotion of the token utilization rate, where we achieve 100\% codebook utilization.
    \item The token sequences used to train the GPT model represent a fixed three-dimensional context of time x height x width of precipitation patterns. This allows the model to learn spatiotemporal dynamics of the evolution of radar sequences.
\end{itemize}

We describe the details of the model setup and novel contributions in the following subsections.

\subsection{Spatial tokenizer: \tokname{}}\label{sec:vqgan}
\begin{figure}[!ht]
	\centering
	\fbox{\includegraphics[width=15cm]{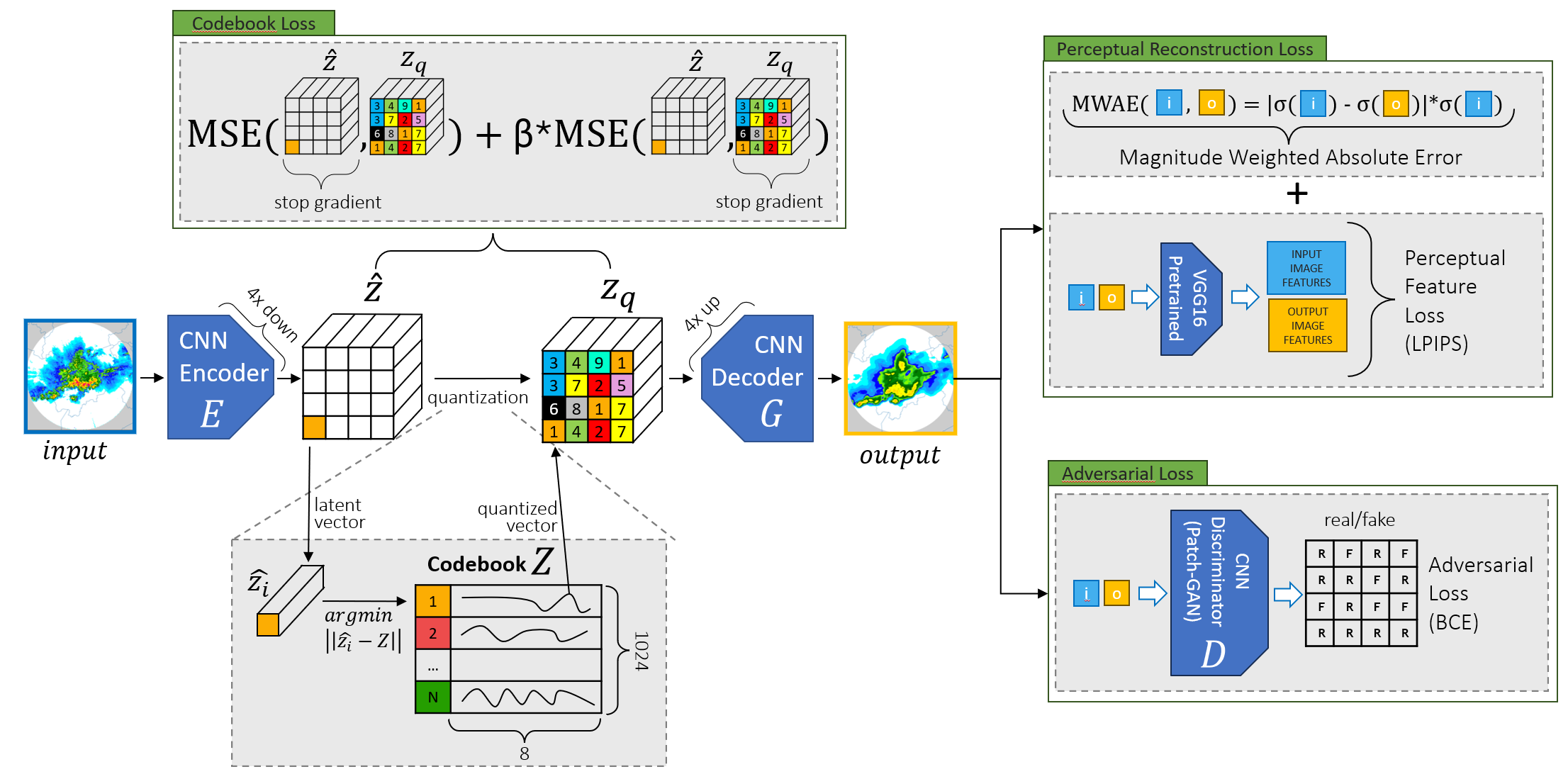}}
	\caption{The spatial tokenizer architecture. The three loss terms are enclosed in boxes with green borders. The blue square [$i$] is the input image, the yellow square [$o$] is the reconstructed autoencoder output.} 
	\label{fig:gptcast_vqvae}
\end{figure}

The spatial tokenizer is a Variational Quantized Autoencoder featuring an adversarial loss (\tokname{})\cite{esser2021taming} and a novel reconstruction loss specifically tailored to improve the reconstruction of precipitation. We carefully tune the architecture of the \tokname{} to obtain a model that provides the highest possible compression, while maintaining a good reconstruction performance and computational complexity. The architecture of the tokenizer is visually summarized in Figure \ref{fig:gptcast_vqvae}.

The encoder ($E$) and decoder ($G$) of the autoencoder are symmetric in design and formed mainly by convolutional blocks, with $\alpha = 4$ steps of downsampling and upsampling, respectively. With this setup, each latent vector at the bottleneck summarises a patch of $2^{\alpha}=2^4=$16x16 pixels of the input image. Following recent studies\cite{yu2022vectorquantized}, we find useful to set a number of channels at the bottleneck (i.e., the length of the latent vector) of 8 to obtain efficient utilization of the codebook, good training stability and the effective capture of essential features in a reduced-dimensional space. The latent vectors at the bottleneck are discretized using a quantization layer that maps them to a finite codebook ($Z$) by finding the closest vector in the codebook. We define a codebook size of 1024 tokens in the quantization layer. The codebook vectors are initialized randomly and then learned during training.

As an example, with an input precipitation map of 192x192 pixels with a dynamic range of 601 possible values for each pixel (from 0 to 60dBZ with a 0.1dBZ step, as described later in Table \ref{table:dataset_summary}), the resulting feature vector at the bottleneck will have a dimensionality of 12H x 12W x 8 channels. Each 8-channel vector is then mapped to one of the possible 1024 vectors in the codebook, resulting in a compressed and discretized representation of 12H x 12W with a dynamic range of 1024 values. The resulting total compression ratio of the spatial tokenizer is $\frac{192 \cdot 192 \cdot 601}{12 \cdot 12 \cdot 1024} \approx 150$ times.

To support such a high compression ratio while maintaining good reconstruction ability, especially for the extreme values, we developed a novel reconstruction loss that we use in place of the commonly used reconstruction losses ($l_{1}$ or $l_{2}$, a.k.a. Mean Absolute Error or Mean Squared Error), defined with the following equation (\ref{eq:mwae}):
\begin{equation}
    \text{MWAE}(\mathbf{x}, \mathbf{y}) = \sum_{i=1}^{n} \left| \sigma(x_i) - \sigma(y_i) \right| \cdot \sigma(x_i) \label{eq:mwae}
\end{equation}

where $\sigma$ is the sigmoid function $\sigma(z) = \frac{1}{1 + e^{-z}}$ and $x$ and $y$ are the input and output vectors of the autoencoder, respectively. We call this loss Magnitude Weighted Absolute Error (MWAE). By giving more weight to pixels with higher rain rates (magnitude), the loss simultaneously serves two purposes: the first is to nudge the tokenizer towards reserving more learning capacity for the reconstruction of extremes, and the other is to try to rebalance the notoriously skewed distribution of precipitation data, that by nature leans towards low rain rates.
We tried different formulations of the loss and found that the introduction of the $\sigma$ function over the inputs improved model convergence, training stability, and codebook usage. The interactions between loss terms during training follow the original \tokname{} implementation\cite{esser2021taming}. The total size of the \tokname{} model is 90M trainable parameters.

\subsection{Spatiotemporal forecaster: GPT}
Similarly to \citeauthor{esser2021taming} the second-stage model is a causal transformer, for our use case we choose a vanilla GPT-2 architecture with 304M parameters.
We train two configurations, one with a spatiotemporal context size of 8 timesteps (40 minutes) x 256 x 256 pixels and a second configuration with 8 timesteps x 128 x 128 pixels. At the token level the two configurations amount to a context length of 2048 (8 x 16 x 16 tokens) and 512 (8 x 8 x 8 tokens) respectively. We refer to the two models as \modnamelarge{} and \modnamesmall{} respectively. In a GPT-like Transformer model, the context size (or sequence length) does not affect the number of parameters, instead, it influences the computational complexity and memory requirements of the model during training and (more crucially) inference. For these reasons, careful considerations in balancing computational complexity and model performance should be made, since timely forecasts are crucial for nowcasting. A summary of the two GPT models' settings is reported in Table \ref{table:gpt_config}.

\begin{table}[h!]
\centering
\caption{\modname{} Model Configurations with large and small spatial domain}
\begin{tabular}{|p{5cm}|p{5cm}|p{5cm}|}
\toprule
\textbf{Configuration/Model name} & \textbf{\modnamelarge} & \textbf{\modnamesmall} \\
\midrule
\textbf{Vocabulary Size} & 1024 & 1024 \\
\textbf{Context Length} & 2048 (8T x 16H x 16W tokens) & 512 (8T x 8H x 8W tokens) \\
\textbf{Number of Layers} & 24 & 24 \\
\textbf{Number of Heads} & 16 & 16 \\
\textbf{Embedding Dimension} & 1024 & 1024 \\
\bottomrule
\end{tabular}
\label{table:gpt_config}
\end{table}

\begin{figure}[!ht]
    \centering
    \fbox{\includegraphics[width=15cm]{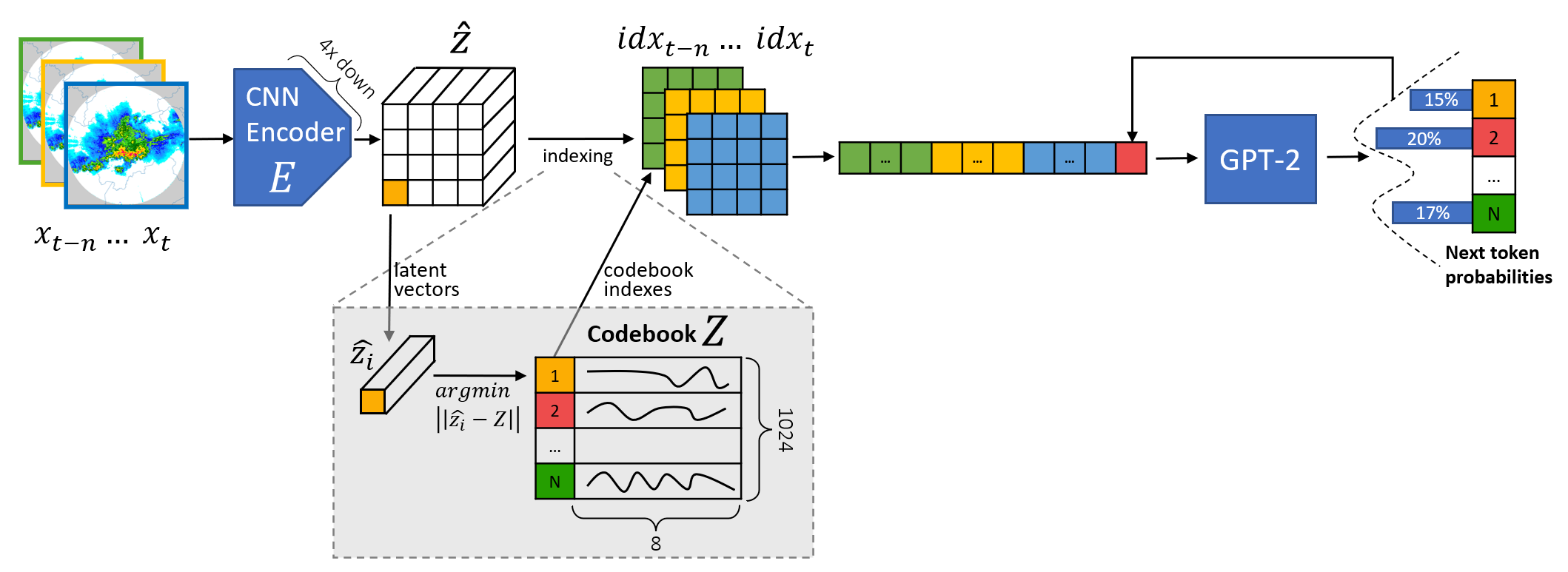}}
    \caption{The spatiotemporal forecaster architecture. During the training of the forecaster the tokenizer encoder ($E$) weights are frozen.}
    \label{fig:gptcast_gpt}
\end{figure}

The training process of the forecaster is schematized in Figure \ref{fig:gptcast_gpt}: contiguous spatiotemporal sequences of radar data are retrieved from the training dataset, and encoded into codebook indices through the frozen \tokname{} encoder and passed to the GPT model as training samples. The indices are ordered starting with the oldest image using a row-first format. The ordering is instrumental to the nowcasting task: in inference, we can provide the model a context that is pre-filled with the past 7 time steps to generate the tokens for the 8th time step. We can generate forecasts for domains with arbitrary sizes by applying a sliding window approach, where we slide the context size across our forecasting domain to predict a target token in the larger domain (starting with the token at the top left position).

\begin{figure}[h!]
	\centering
	\fbox{\includegraphics[width=15cm]{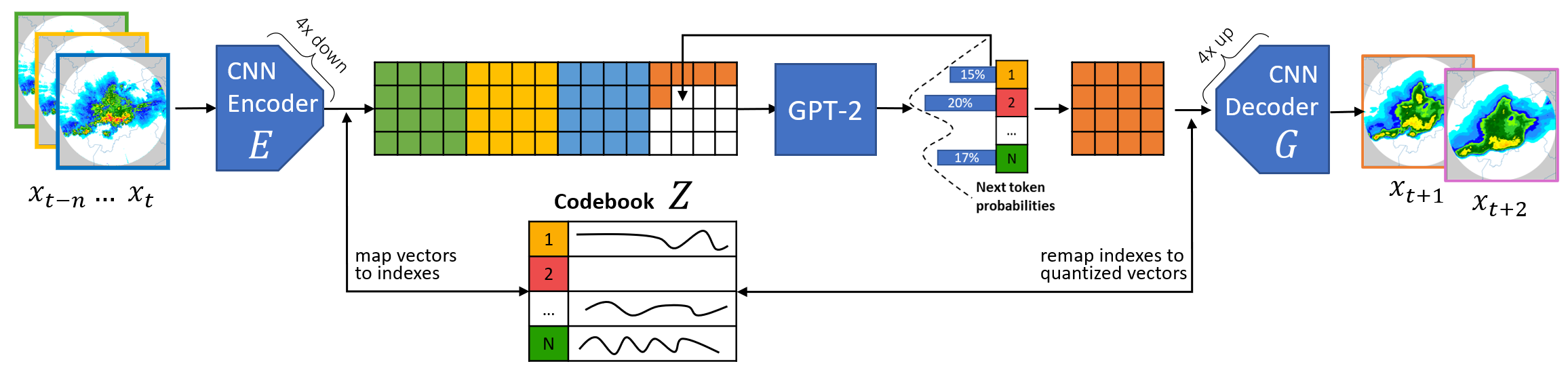}}
	\caption{The \modname{} architecture during inference. The trained tokenizer and forecaster are combined (Tokenizer encoder ($E$) -> Forecaster -> Tokenizer decoder ($G$)) to generate forecasts. In the standard unconditional setting, the next token is chosen by applying a multinomial draw over the codebook probabilities to generate different ensemble members.}
	\label{fig:gptcast_inference}
\end{figure}

At inference time the two models are combined in a sandwich-like configuration, with the encoding of the context input images through the \tokname{} encoder, the autoregressive generation of the indices of multiple forecasts steps via the transformer model, and the final decoding of the tokens back to pixel space using the \tokname{} decoder (see Figure \ref{fig:gptcast_inference}). To obtain multiple ensemble members, the autoregressive generation of the indices can be repeated multiple times while applying a multinomial draw over the output probabilities to pick different tokens.

\section{Dataset}
\begin{figure}[h!]
	\centering
	\includegraphics[width=12cm]{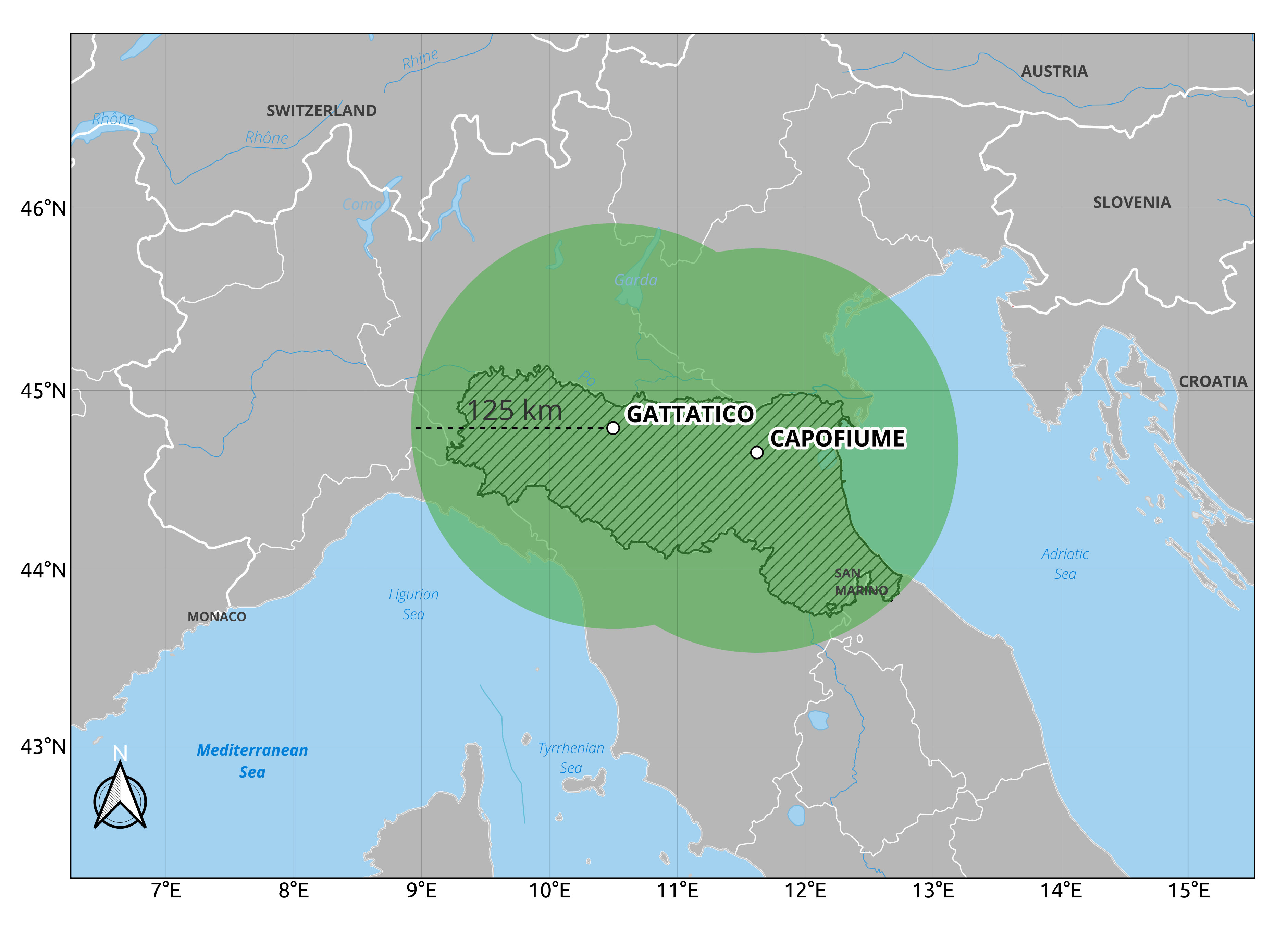}
	\caption{Extent of the dataset. Effective coverage is the composite of the 125 km range of the Gattatico and San Pietro Capofiume radars (green area). Hatched area is the Emilia-Romagna Region.}
	\label{fig:workspace}
\end{figure}

The dataset we propose for the study is the radar reflectivity composite produced by the HydroMeteorological Service of the Regional Agency for the Environment and Energy of Emilia-Romagna Region in Northern Italy (Arpae Emilia-Romagna). The agency operates two Dual-polarization C-Band radars in the area of the Po Valley, located  respectively in Gattatico (44°47'27"N, 10°29'54"E) and San Pietro Capofiume (44°39'19"N, 11°37'23"E). The scanning strategy allows coverage of the entire Region every 5 minutes. The area is characterized by a complex morphology and it spans from the flat basin of the Po valley in the north to the upper Apennines in the south, and from the Ligurian coast in the west to the Adriatic Sea in the east. For the purpose of this work, scans with a radius of 125 km were chosen with a total coverage of 71172 square km, summarized in Figure \ref{fig:workspace}.

Arpae fully manages both the radar acquisition strategy and the data processing pipeline. They include several stages of data quality control and error correction developed to reduce the effect of topographical beam blockage, ground clutter, and anomalous propagations \cite{fornasieroEnhancedRadarPrecipitation2006}. Specific corrections are applied over the vertical reflectivity profile to improve precipitation estimates at the ground level\cite{fornasieroRadarQuantitativePrecipitation2008}.

The resulting product used for this study is a 2D reflectivity composite map on a 290 x 373 km grid at 1km resolution per pixel, with a time step of 5 minutes. Reflectivity values range from -20dBZ to 60 dBZ. When converting reflectivity values to rain-rate (mm/h) the standard Marshall-Palmer Z-R relationship with  $a = 200$ and $b = 1.6$ is applied  \cite{marshall1948distribution}.

\subsection{Data selection, preprocessing and augmentation}

For the purpose of our study, we extract all contiguous precipitating sequences in the 6 years between 2015 and 2020. Non-precipitating sequences are discarded, resulting in the selection of 179,264 timesteps out of 630,720 ($71,5\%$ of the data is discarded). The precipitating sequences are divided between training, validation, and test sets.

We prepare two test sets, one for the testing of the spatial tokenizer and one for the testing of the forecaster. To test the spatial tokenizer we isolate all time steps belonging to the days in the years 2019 and 2020 where extreme events happened by analyzing historical weather reports, resulting in a total of 21,871 radar images (time steps). We call this the \emph{Tokenizer Test Set} (TTS). To test the forecaster we follow the same validation approach of \citeauthor{pysteps}, and we extract out of the  \emph{Tokenizer Test Set} 10 sequences of 12 hours each representative of the most relevant events. This 120-hour subset, namely the \emph{Forecaster Test Set} (FTS), is used for the testing of the forecaster.

The remaining sequences are randomly divided between training and validation, with the following final result: 149,524 steps for training, 7,869 for validation, 21,871 for the TTS that includes 1450 steps (12 hours * 10 events) of the FTS. To further increase the training dataset size and promote generalization we apply random cropping, random 90-degree rotation and flipping to the training dataset during the training phase. The data values are preprocessed by clipping the reflectivity range between 0 and 60 dBZ to minimize the contribution of spurious echoes and drizzle, and by rounding the values to the first decimal digit, resulting in an effective dynamic range of 601 values (from 0 to 60 with a 0.1 step) per pixel.

Table \ref{table:dataset_summary} summarizes the resulting dataset characteristics.

\begin{table}[h!]
\centering
\caption{Summary of dataset characteristics}
\begin{tabular}{|p{3cm}|p{11cm}|}
\toprule
\textbf{Attribute} & \textbf{Details} \\
\midrule
\textbf{Product Description} & Arpae radar reflectivity composite (central Italy) \\
\textbf{Map Size} & 290 x 373 pixels \\ 
\textbf{Pixel Size} & 1km resolution \\ 
\textbf{Timestep} & 5 minutes \\ 
\textbf{Reflectivity Range} & -20–60 dBZ (clipped to 0-60 dBZ, 0.1 step = 601 values of dynamic range) \\ 
\textbf{Date range} & precipitation sequences in the years 2015 - 2020 \\ 
\textbf{Dataset size} & 630,720 total timesteps (179,264 timesteps selected)  \\
\textbf{Train and validation} & 149,524 timesteps for training, 7,869 validation \\
\textbf{Test datasets} & \emph{TTS}: 21,871 timesteps, \emph{FTS}: 1450 timesteps (10 events of 12 hours) \\
\bottomrule
\end{tabular}
\label{table:dataset_summary}
\end{table}

\section{Results}
We analyze the performances of our model at two stages: first, we analyze the amount of information loss introduced by the data compression in the tokenizer, and then we analyze the performance of \modname{} as a whole for the nowcasting of precipitation up to two hours in the future.

\subsection{Spatial tokenizer reconstruction performances}
Given the high compression ratio that we introduce in the \tokname{} it is crucial to understand how much and what type of information is lost during the compression and discretization step operated by the tokenizer. Depending on the nature of the information loss, certain phenomena may be completely lost and this can compromise the ability of the transformer to learn and forecast some precipitation dynamics (e.g. extreme events).
The new MWAE loss introduced in Section \ref{sec:vqgan} is specifically built to improve the reconstruction performances of the tokenizer and reach a good level of data reconstruction while maintaining a high compression factor.

Table \ref{table:result_vqvae} shows the performances in reconstruction ability on the TTS between a \tokname{} trained using as reconstruction loss a standard Mean Absolute Error (MAE) and using our proposed MWAE loss. We consider both global regression scores like Mean Absolute Error (MAE), Mean Squared Error (MSE), and the Structural Similarity Index Measure (SSIM, \cite{wang2004image}) along with categorical scores computed by thresholding the precipitation at multiple rain rates (1, 10 and 50 mm/h), like the Critical Success Index (CSI) and the frequency bias (BIAS).

\begin{table}[h!]
\centering
\caption{Reconstruction performance on the TTS of \tokname{} trained with Mean Absolute Error (MAE) loss and with our proposed MWAE loss. $(\downarrow)$ means lower is better, $(\uparrow)$ means higher is better, and for frequency bias (BIAS) closer to 1 is better.}
\begin{tabular}{|p{2.6cm}|>{\raggedright\arraybackslash}p{1.4cm}|>{\raggedright\arraybackslash}p{1.4cm}|>{\raggedright\arraybackslash}p{1.4cm}|>{\raggedright\arraybackslash}p{2.3cm}|>{\raggedright\arraybackslash}p{2.0cm}|>{\raggedright\arraybackslash}p{2.0cm}|}
\toprule
model / performance & \textbf{MAE} $(\downarrow)$ & \textbf{MSE} $(\downarrow)$ & \textbf{SSIM} $(\uparrow)$
& \textbf{CSI $(\uparrow)$ / BIAS @ 1 $mm^{-h}$} & \textbf{CSI / BIAS @ 10 $mm^{-h}$} & \textbf{CSI / BIAS @ 50 $mm^{-h}$} \\
\midrule
\textbf{\tokname{} MWAE} & 0.204 & 4.09 & 0.988 & 0.81 / 1.03 & 0.56 / 0.94 & 0.44 / 0.92 \\
\textbf{\tokname{} MAE} & 0.265 & 7.09 & 0.981 & 0.74 / 0.93 & 0.38 / 0.62 & 0.13 / 0.22 \\
\bottomrule
\end{tabular}
\label{table:result_vqvae}
\end{table}

The autoencoder trained with MWAE shows significant improvements over all the considered metrics, but it is crucial to notice that the improvements are more pronounced for higher rain rates, whose frequency is almost precisely reconstructed by the autoencoder. This is clearly visible in the improvements in BIAS at 50mm/h, which is defined as the fraction between the number of pixels in the input image over 50 mm/h and the number of pixels that surpass the same threshold in the reconstruction, where we obtain a jump in performance from 0.22 to 0.92 (where 0 is total underestimation, 1 is the perfect score, and greater than 1 is overestimation).

\begin{figure}[!ht]
    \centering
    \includegraphics[width=17cm]{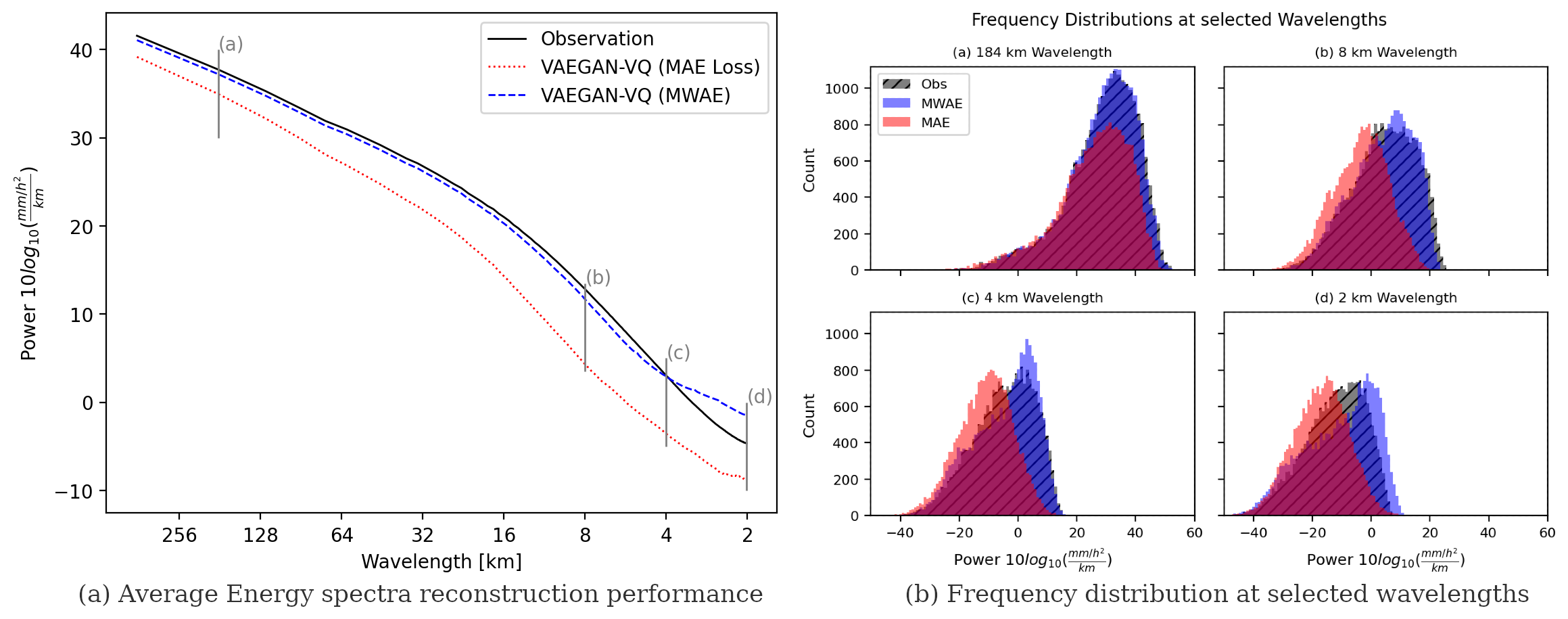}
    \caption{Comparison of radially averaged power spectral density reconstruction performance by adopting the MWAE loss function compared to MAE. The adoption of MWAE improves the ability of the autoencoder to reproduce the energy distribution of precipitation at all wavelengths.}
    \label{fig:power}
\end{figure}

The recovery in frequency is also confirmed by analyzing the radially averaged power spectral density (i.e., the amount of energy) of the input and reconstruction: as shown in Figure \ref{fig:power}, the average power spectra of the MWAE autoencoder closely resembles the input (albeit with an overestimation at the smallest wavelengths), while the standard autoencoder distribution is constantly shifted and underestimated at all wavelengths.

Improvement in CSI score is also significant (at 50 mm/h, more than three times higher), albeit not as thorough as the frequency recovery. This implies that the remaining source of error is that the reconstructed precipitation fields have either a different structure or a different location when compared to the input (i.e., the amounts of the reconstructed precipitation are correct but misplaced at the spatial level).

To better characterize this remaining source of error, we compute the SAL measure \cite{wernli2008sal, wernli2009spatial}, which evaluates three key aspects of the precipitation field within a specified domain: structure (S), amplitude (A), and location (L). The amplitude component (A) measures the relative deviation of the domain-averaged reconstructed precipitation amount from the input. Positive values indicate an overestimation of total precipitation, while negative values indicate an underestimation. The structure component (S) assesses the shape and size of predicted precipitation areas. Positive values occur when these areas are too large or too flat, while negative values indicate that they are too small or too peaked. The location component (L) evaluates the accuracy of the predicted location of precipitation. It combines information about the displacement of the reconstructed precipitation field’s center of mass compared to the input and the error in the weighted average distance of the precipitation objects from the center of the total field. Perfect forecasts result in zero values for all three components, indicating no deviation between input and reconstructed precipitation patterns.

\begin{figure}[!ht]
\centering
\includegraphics[width=17cm]{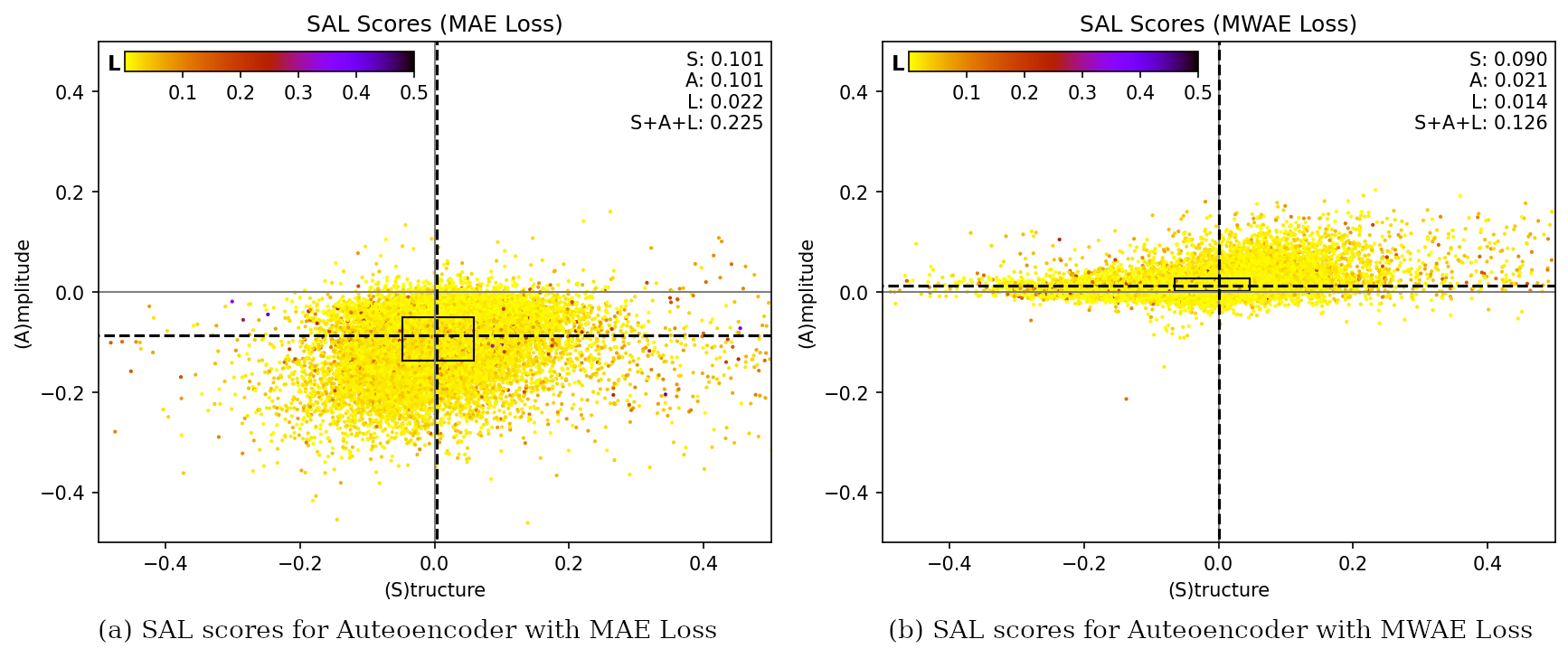}
\caption{Structure, Amplitude, and Location (SAL) plot that compares the performance of the MAE and MWAE autoencoders. Each dot on the plot represents the scores of one image in the TTS. Structure and amplitude are plotted on the horizontal and vertical axes, respectively, while the location component is represented by the color. The dashed vertical and horizontal lines indicate the median values of the Structure (S) and Amplitude (A) scores, respectively. The rectangle box represents the area between the 25th and 75th percentiles (i.e., 50\% of the dots fall inside the boxed area). The numbers on the top right show the Mean Absolute values.}
\label{fig:sal}
\end{figure}

The SAL analysis plot for both autoencoders is shown in Figure \ref{fig:sal}. The MWAE autoencoder improves over the baseline autoencoder on all scores, with a median value that is close to zero for all three components. A residual source of absolute error remains in the Structure component, while both Amplitude and Location errors are negligible.

\begin{figure}[!ht]
\centering
\includegraphics[width=14cm]{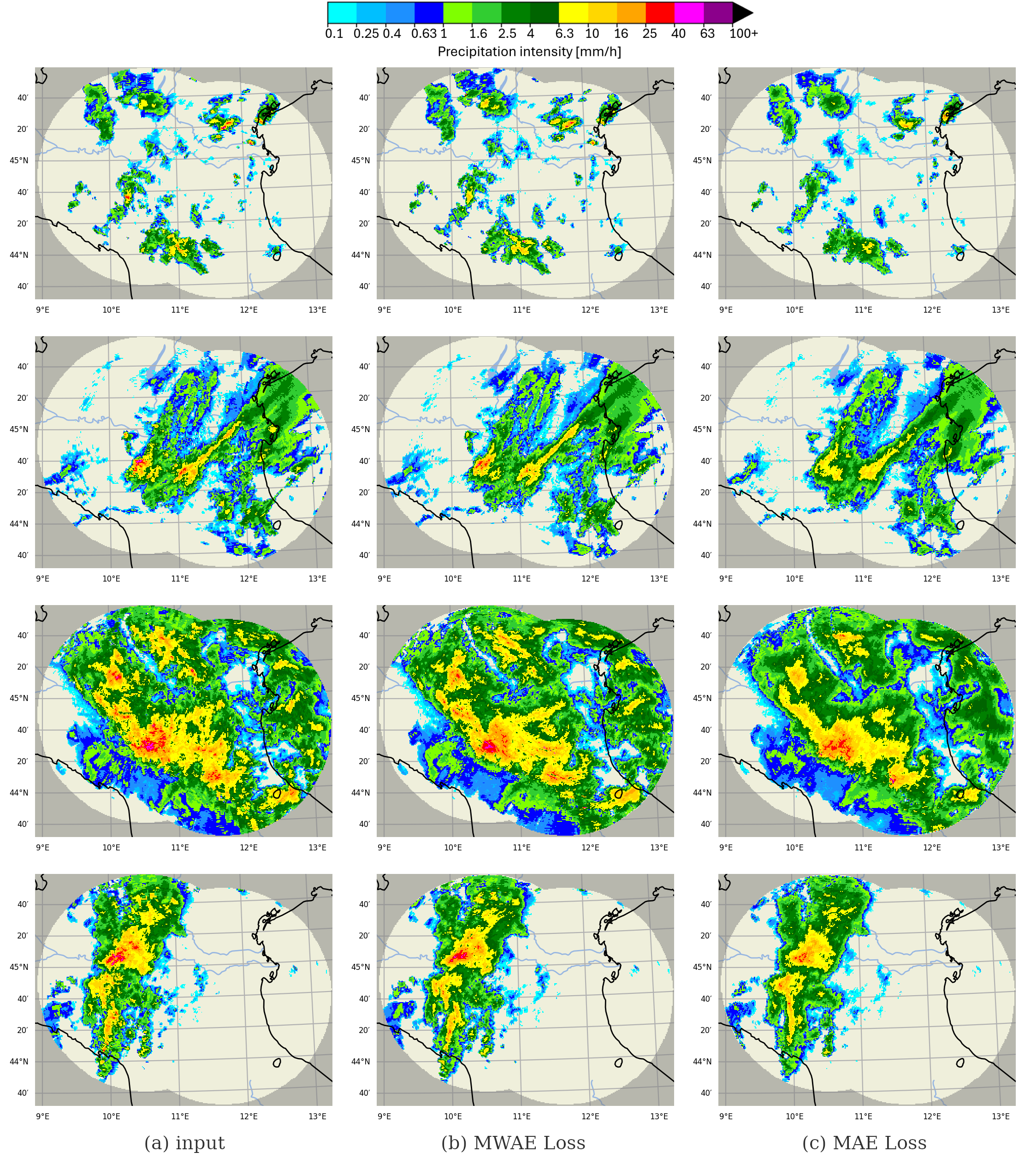}
\caption{Qualitative comparison between precipitation snapshots reconstructed by the \tokname{} autoencoder trained with MWAE loss and MAE loss, taken from the TTS. The autoencoder trained with MWAE loss shows a marked improvement in the reconstruction of precipitation, with crucial improvements in the reconstruction of higher rain rates (thunderstorms). }
\label{fig:vae_comparison}
\end{figure}

In summary, divergences in the size and shape of the reconstructed precipitation patterns account for the majority of the error for our new autoencoder, while the locations, frequencies, and energy contents of the precipitation patches are mostly accurate. Overall, this is a good compromise for the nowcasting task since we can tolerate higher compromises for errors in structure, whereas systematic errors in amplitude, frequency, or location can seriously impair the forecaster's ability to accurately predict the evolutionary dynamics of precipitation. Some qualitative examples of the input and reconstruction from both autoencoders are presented in Figure \ref{fig:vae_comparison}.

\subsection{\modname{} Nowcasting performances}
We examine and compare \modname{} forecasting performance with that of the Lagrangian INtegro-Difference equation model with Autoregression (LINDA)\cite{LagrangianIntegroDifferenceEquationModelforPrecipitationNowcasting}, the state-of-the-art ensemble nowcasting model included in the pySTEPS package\cite{pysteps}. LINDA is a nowcasting technique intended to provide superior forecast skill in situations with intense localized rainfall compared to other extrapolation methods (S-PROG or STEPS). Extrapolation, S-PROG\cite{sprog}, STEPS\cite{steps}, ANVIL\cite{anvil}, integro-difference equation (IDE), and cell tracking techniques\cite{titan} are all combined in this model.

For the comparison, we use the FTS. Out of the 10 events in FTS, 7 are convective events occurring in spring or summer, and three are winter precipitation events. For each event, we produce a forecast every 30 minutes, and each forecast is a 20-member ensemble forecast with 5-minute time steps and a maximum lead time of 2 hours (i.e., 24 forecasting steps) for both LINDA and \modname{}. This results in a total of 200 forecasts (20 forecasts per event) generated per model.
For \modname{} we test both the two model configurations, \modnamelarge{} and \modnamesmall{}.

For verification assessment, we rely on the Continuous Ranked Probability Score (CRPS) and the rank histogram, which are essential tools for verifying ensemble forecasts. By showing the frequency of observed values among the forecast ranks, the rank histogram evaluates the dispersion and reliability of ensemble forecasts and highlights biases such as under- or over-dispersion. By comparing the prediction's cumulative distribution function to the actual value, CRPS calculates a numerical score for forecast skill that indicates how accurate a probabilistic forecast is. The two scores complement each other, with the CRPS providing a measure of forecast accuracy as a whole and the rank histogram emphasizing the ensemble spread and reliability.

\begin{figure}[h!]
	\centering
	\fbox{\includegraphics[width=13cm]{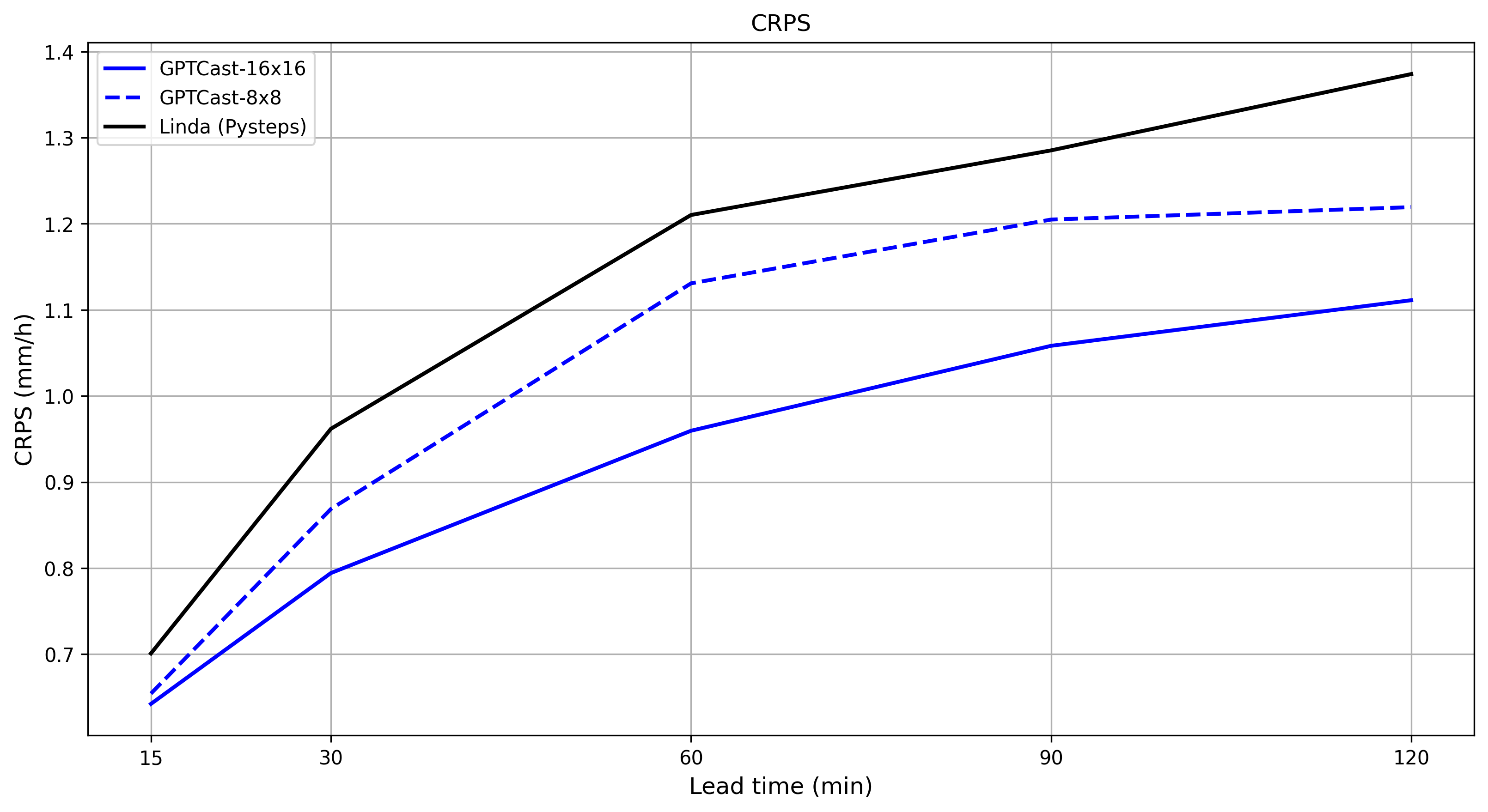}}
	\caption{CRPS (continuous ranked probability score) comparison of \modname{} and LINDA over the FTS (lower is better) at different lead times.}
	\label{fig:crps}
\end{figure}

The CRPS score for each of the three models—LINDA, \modnamelarge{}, and \modnamesmall{}—is displayed in Figure \ref{fig:crps}: both variants of \modname{} outperform LINDA across all lead times, with \modnamelarge{} outperforming all other models. This result clearly shows that the model can learn a more thorough dynamic of the evolution of precipitation patterns when the context size is more spatially extended. It is important to notice that this improvement comes with a non-negligible increase in terms of computational time at inference, which in our experiments was close to an order of magnitude (\modnamesmall{} computes a timestep in 2 seconds compared to 17 seconds for the larger model on an NVIDIA RTX 4090).

\begin{figure}[!ht]
\centering
\includegraphics[width=.8\linewidth]{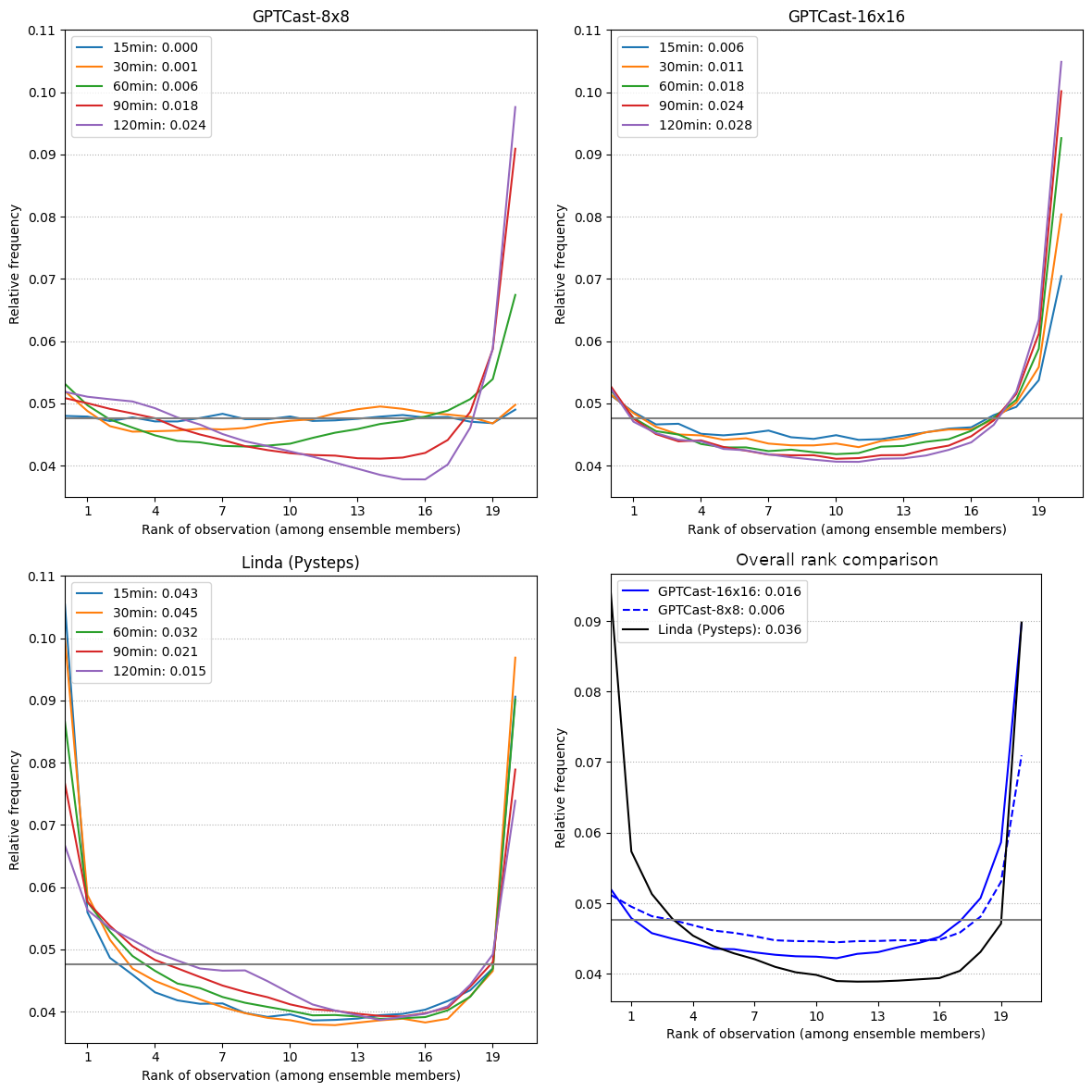}
\caption{Rank histrograms comparison of \modname{} and LINDA on the FTS. The horizontal gray line represents the ideal value (the closer the better). The numbers in the legend indicate the Kullback–Leibler divergence from the uniform distribution (lower is better). }
\label{fig:rankhist}
\end{figure}

Figure \ref{fig:rankhist} analyzes the rank histogram at different lead times for all three models, including information on the Kullback–Leibler divergence (KL) from the uniform distribution. Both versions of \modname{} provide a better overall score over LINDA that tends to be under-dispersed, with \modnamesmall{} being the best model. Moreover, \modnamesmall{} shows a rank distribution close to optimal up to the first hour, with a KL divergence from the uniform distribution of 0.006 at 60 minutes lead time (12 steps). \modnamelarge{} displays an overall better rank histogram than LINDA up to the first 60 minutes with a tendency to underestimation that compounds over time: we attribute this behavior to the increased ability of the \modnamelarge{} to capture the training distribution, that has a higher ratio of dissipating precipitation events than the FTS (which is filtered to contain only extreme events).

\begin{figure}[!ht]
\centering
\includegraphics[width=.75\linewidth]{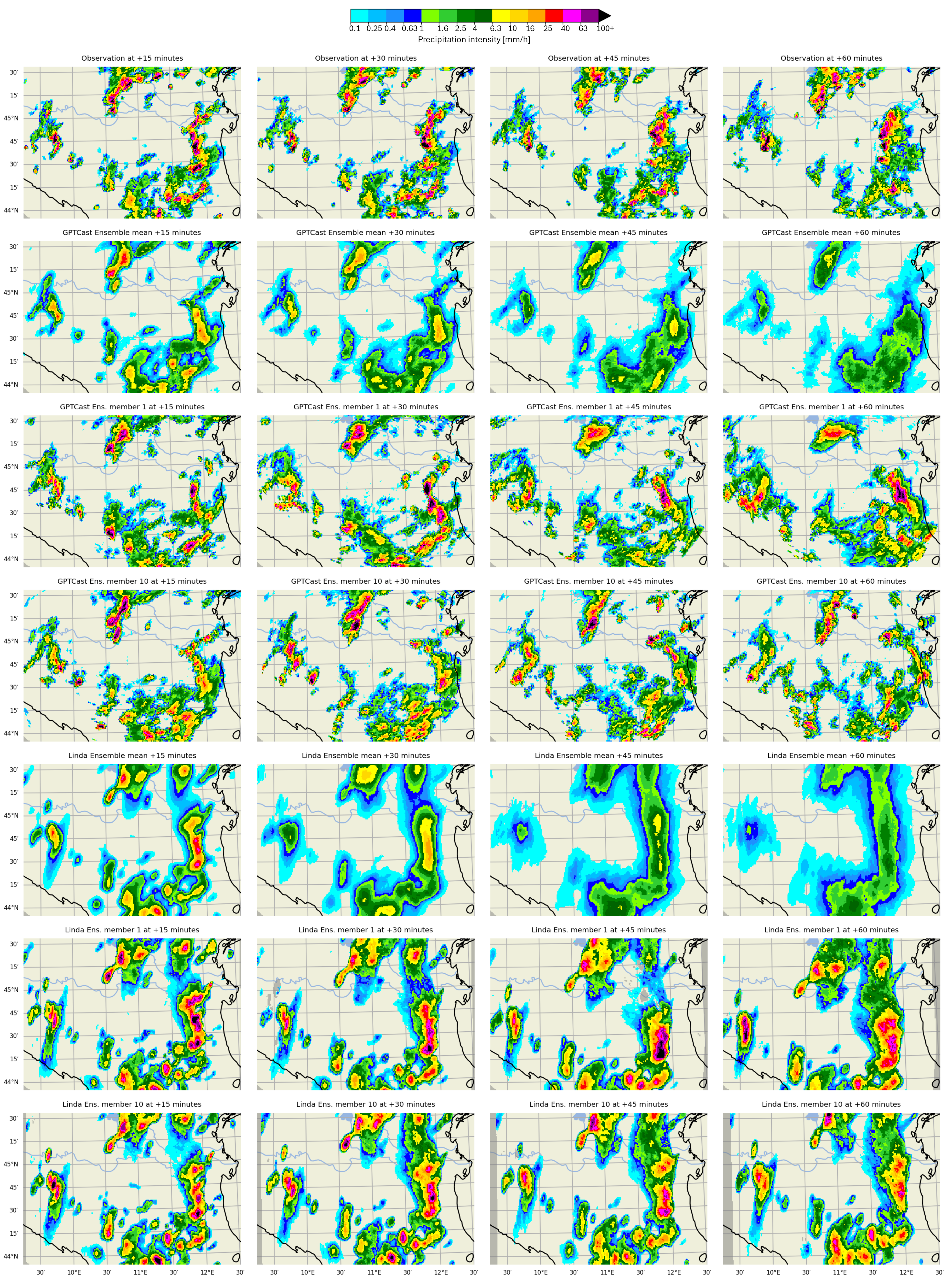}
\caption{Example comparison of \modnamelarge{} and LINDA nowcast on a convective case in the Forecaster Test Set (2020-06-08 11:00 UTC). The domain is cropped on the central area for visualization convenience.}
\label{fig:verif}
\end{figure}

Figure \ref{fig:verif} shows an example of nowcast for a convective case in the FTS, with two ensemble members and the ensemble mean for both LINDA and \modname{}. \modname{} generates two realistic and diverse forecasts, with an ensemble mean that features a better location accuracy than LINDA compared to the observations.

\conclusions[Discussion and future work]  
\modname{} introduces a novel approach to ensemble nowcasting of radar-based precipitation, leveraging a GPT model and a specialized spatial tokenizer to produce realistic and accurate ensemble forecasts. We show that this approach can provide reliable forecasts, outperforming the state-of-the-art extrapolation method in both accuracy and uncertainty estimation. 

\modname{}'s deterministic architecture enhances interpretability and reliability by generating realistic ensemble forecasts without random noise inputs. The model can be declined in different sizes, both in context length and in terms of parameters (which we postponed to future analyses) allowing to balance the trade-off between accuracy and computational demands and providing flexibility for different operational settings.

We believe that our method, by adopting an architecture influenced by large language models (LLMs), paves the way for future promising research in precipitation nowcasting that can incorporate all the improvements and developments from the quickly developing field of LLM research. This includes more efficient architectures, improved training techniques, and better interpretability tools. Such integration can potentially enhance \modname{}'s performance, scalability, and usability, ensuring that it remains a state-of-the-art nowcasting tool.

Despite its strengths, the approach poses specific challenges that must be considered for the operational usage of the model. The approach requires training two models in cascade, each with its own set of challenges. In our experiments, it was hard to find a stable configuration to train the spatial tokenizer that has to balance multiple competing losses. The MWAE reconstruction loss we introduced helped substantially in terms of both convergence and stability, although at the cost of slower training induced by the smoothing effect of the sigmoid ($\sigma$) terms in the loss.
On the other hand, we found the forecaster to be very stable in training (as expected by transformers) but computationally intensive in inference, especially for the long context configuration (\modnamelarge{}), making its use in a real-time application like nowcasting challenging without significant resources.
The ability of the model to effectively capture the training distribution is both its main strength and point of attention.
From an operational perspective, our hypothesis is that, due to the distinct distribution of stratiform and convective precipitation, training separate models for stratiform (winter) and convective (summer) precipitation may result in better forecasts. This implies that a larger and better-quality dataset may be needed than the one used in this work to avoid model overfitting.

Future work could explore optimizing context size and computational complexity to balance performance and resource demands, as well as integrating the vast literature about more efficient transformer architectures (e.g., flash attention, speculative decoding, etc...). We also plan to explore the interpretability of the model to control and condition the model for different tasks. The peculiar characteristics of \modname{} open the possibility of guiding the generative process of the model by combining the probabilistic output of the forecaster with the interpretability of the learned codebook in terms of physical quantities. A possibility that we envision is to leverage \modname{} for tasks like seamless forecasting (a.k.a. blending), generation of what-if scenarios, forecast conditioning, weather generation, and observation correction capabilities.




\codedataavailability{Data from ARPAE Emilia Romagna .
The full, preprocessed dataset used for the presented experiments is available on Zenodo \citep{franch_2024_gptcast_data}, including the generated ensemble forecasts to reproduce the verification scores. The pretrained models are available on Zenodo \citep{franch_2024_gptcast_models}. A dedicated GitHub repository (https://github.com/DSIP-FBK/GPTCast) hosts the Pytorch Lightning \citep{Falcon_PyTorch_Lightning_2019} code of the models described in this paper, based on the Lightning-Hydra-Template \citep{Yadan2019Hydra}, licensed under the MIT License. The repository also hosts the code to reproduce the images shown in this paper. \modname{} v1.0 GitHub release is archived on Zenodo \citep{franch_2024_gptcast_code} and allows to download the code to reproduce the presented experiments.} 












\authorcontribution{GF conceived and conceptualized the study, designed the GPTCast architecture, implemented the code and ran the experiments. GF and ET performed the analysis and verification of the results and wrote the manuscript. VP, CC, PPA provided the data, performed the data extraction, data selection and data quality control. RW performed data format conversion. All authors revised the results and reviewed the manuscript. MC supervised the study from end to end.} 

\competinginterests{The authors declare that they have no conflict of interest.} 


\begin{acknowledgements}
We acknowledge CINECA Consortium for providing the GPU resources for training and running the experiments presented in this study.
\end{acknowledgements}







\bibliographystyle{copernicus}
\bibliography{references.bib}

\end{document}